# Context Aware Query Image Representation for Particular Object Retrieval

Zakaria Laskar*, Juho Kannala

Aalto University
{zakaria.laskar,juho.kannala}@aalto.fi

**Abstract.** The current models of image representation based on Convolutional Neural Networks (CNN) have shown tremendous performance in image retrieval. Such models are inspired by the information flow along the visual pathway in the human visual cortex. We propose that in the field of particular object retrieval, the process of extracting CNN representations from query images with a given region of interest (ROI) can also be modelled by taking inspiration from human vision. Particularly, we show that by making the CNN pay attention on the ROI while extracting query image representation leads to significant improvement over the baseline methods on challenging Oxford5k and Paris6k datasets. Furthermore, we propose an extension to a recently introduced encoding method for CNN representations, regional maximum activations of convolutions (R-MAC). The proposed extension weights the regional representations using a novel saliency measure prior to aggregation. This leads to further improvement in retrieval accuracy.

## 1 Introduction

With the introduction of scale invariant local features, such as SIFT [1], the field of image retrieval has benefited tremendously over the last decade, extending its popularity and applicability to other fields of research like loop closure in robotics [35], [36] and structure from motion [34]. In particular, the extension of bag-of-words model from text retrieval to the case of particular object retrieval in videos by Sivic and Zisserman [2] has been seminal for the developments in image retrieval [3],[4]. One of the advantages of such local features is that it allows to follow-up the initial retrieval results with a costly but more accurate spatial verification step [23]. The initial retrieval results are obtained by matching local descriptors using selective matching kernels [33]. The issue of scalability in terms of memory requirements and computational cost associated with pairwise descriptor matching for large scale image retrieval databases was addressed by encoding the local descriptors into a single compact global image representation. Popular techniques are Fisher vectors [5], and VLAD [6].

---

* Contact Author.

The increase in computational capacity of GPUs and the generation of large datasets, such as ImageNet [10] has made Convolutional Neural Networks (CNN) the popular choice for a broad spectrum of computer vision tasks like image classification [9], object detection [25], scene recognition [26], camera pose estimation [24]. As training a CNN from scratch requires a large amount of data, using activations from different layers of a CNN, trained on a large dataset like [10], as off-the-shelf image representation has bridged the applicability of CNN to different domains [11], [12], [13]. In case of limited amount of data, the parameters of a CNN pre-trained on ImageNet or other large datasets can be used to initialize the network parameters before training the CNN on the target dataset. Such a process is known as fine-tuning and has leveraged the use of CNN to other domains [17], [18]. For the case of image retrieval, several works [8], [7], [13], [16] propose the use of activations from a pre-trained CNN as image descriptors. As the pre-trained CNNs employed for such instance-level retrieval tasks were generally trained to suppress intra-class variations (observed in generic computer vision problems, like object detection), the performance of CNN based descriptors lagged behind the conventional local descriptors. Babenko et al. [7] first demonstrated that fine-tuning a pre-trained CNN on a Landmark dataset [7] significantly improves the retrieval accuracy on standard benchmark datasets of landmarks, such as Oxford5k [3] and Paris6k [18]. Arandjelovic et al. [19] used a similar paradigm of learning the image representations from a large dataset of geo-tagged images. However, the training was done using a ranking loss instead of the classification loss used in [7]. Radenovic et. al. [43] showed the importance of hard positive and hard negative mining using unsupervised methods in improving the retrieval accuracy. Post training, the final representation of an image is encoded/computed using regional maximum activations of convolutions (R-MAC) [23], [43]. R-MAC aggregates maximum activations over multiple regions into a compact image representation. The regions are generated using a fixed grid, which is designed to make the final image representation robust to scale and translation variation. Gordo et al. [20] proposed learning R-MAC representation using a Region Proposal Network (RPN) [28]. The R-MAC and the RPN was trained in an end-to-end manner resulting in powerful image representation that obtain the existing state-of-the-art performance on benchmark retrieval datasets.

In this paper, we focus on particular object retrieval, which is a special case of image retrieval, whereby, a query image(s) is given along with a region of interest (ROI) containing an object of interest. The retrieval engine then returns a ranked list of the database images, such that the images containing the object of interest are ranked higher. Traditionally, image retrieval methods encode the query image using feature representation extracted only from the ROI. This serves two purposes : reduced interference from background clutter, and, suppression of distractive patterns outside the ROI, which, sometimes maybe more salient than the ROI. On the other hand, regions outside the ROI can add contextual information to the ROI representation that can facilitate improvement in retrieval performance. Thus it can be stated that the suppression and encoding of information from regions outside the ROI is a tightly coupled problem. To address this issue, we propose the extension of computational model of hippocampus spatial attention, introduced by Mozer et al. in 1998 [38], to the problem of particular object retrieval. In particular, we show that by partial suppression of intermediate CNN representations (representations from intermediate layers of the CNN) of regions in the query image outside the ROI, we can obtain a proper trade-off between the two problems stated above. The related work by Chum et al. [4] also address this issue by selectively growing the query model beyond the ROI using the images from the initial retrieval results. As such their model can only be applied as a post-processing step on top of the results obtained from the initial query made by using only the ROI representation. On the other hand, we propose to encode the context along with the ROI in a single global image representation. Our results demonstrate the effectiveness of such context aware query image representation in retrieval tasks on standard benchmark datasets like Oxford [3] and Paris [18].

Our second contribution is that we propose an extension to the conventional R-MAC encoding technique. The standard R-MAC suffers from the drawback of assigning uniform weightage to all the regions generated by the pre-defined grid prior to aggregation. As the regions are generated independent of the image content, responses from background clutter can cause negative interference due to equal weightage. Gordo et al. [20] proposed the use of RPN to generate image content dependent regions to circumvent this problem. However, this has certain challenges : i) the number of regions from RPN is 3 to 10 times more than R-MAC [37], and, ii) additionally, the RPN model parameters need to be trained for the given task separately. Instead, we show that by using a simple saliency measure, obtained from the existing representation technique, one can weight the regional representations of R-MAC before aggregating. The saliency measure assigns higher weight to landmark type regions and lower weight to the background clutter.

Using the proposed modifications, we are able to achieve state-of-the-art retrieval results in standard object retrieval datasets. Our work can be seen as an extension to the work of [20], [37] since we use off the shelf CNN representations from their trained network.

## 2   Background

In this section, we provide the reader with a brief background with the methods and terminologies encountered in CNN based literature.

### 2.1   CNN

When using a pre-trained CNN network like VGGNet (VGG) [29] or Residual Network (ResNet) [30], the network is often cropped at the last convolutional or pooling layer [18], [20]. For example, `conv5/pool5` layer in VGG, or `res5c` in Resid-101. As such in the remainder of the paper, the term CNN will be associated with such cropped networks. Now, consider a CNN with $L$ layers. Given an input image $I \in \mathbb{R}^{W_I \times H_I \times 3}$, the response obtained at the output of the layer $l \in L$ is a 3D tensor $\boldsymbol{X}^l \in \mathbb{R}^{W \times H \times K}$. $K$ is the number of channels and $W \times H$ the spatial dimensions of the output feature map. The spatial resolution of the feature map depends on the network architecture and size of the input image, while the number of output channels $K$ equals the number of filters in layer $l$. Additionally, it is assumed the feature map $\boldsymbol{X}^l$ is passed through a rectified linear unit (ReLU) activation function to ensure the non-negativeness of the activations.

The feature map $\boldsymbol{X}^l$ can be denoted as a set of $K$ 2D feature maps $\boldsymbol{X}^l = \{X_k^l\}, k = 1...K$. Alternatively, each feature map $\boldsymbol{X}^l$ can be said to be a set of $K$ dimensional feature representations for $W \times H$ activations. Instead of the term 'pixels', the term 'activation' is used in CNN feature space. The activation at spatial location $p \in \mathbb{R}^2$ in feature map $X_k^l$ is represented by $X_{k,p}^l$. The set of all such locations $p$ in a feature map be represented by $\boldsymbol{S} = [1, W] \times [1, H]$. As the layers are arranged in an hierarchical order, each layer computes a higher level abstraction from the feature representations of the previous layers.

## 2.2  R-MAC

Although the feature maps represent high quality abstraction of the image, they are very high dimensional. Typical dimensionality of feature maps extracted at the output of `res5c` layer in ResNet-101 network is $23 \times 13 \times 2048$ for an image of spatial resolution $800 \times 600$. As such, these high dimensional representations are encoded to fixed length global representations using techniques [19] , [14], [15] and [16]. Among the various encoding methods, R-MAC has shown highest performance [16] and is a widely popular choice of encoding CNN representations.

As R-MAC can be used to encode feature map from any layer, $l$, we continue with the same set of notations introduced in Sec.2.1, and, do not introduce any layer specific notations. For a given feature map, $\boldsymbol{X}^l$, R-MAC first generates a set of rectangular regions $\boldsymbol{R} = \{R_i\}, i = 1...N$, where $R_i \subseteq \boldsymbol{S}$ and $N$ is the number of regions that depends on the size of feature map. For each region, the maximum activations of convolutions (MAC) [8] is computed by spatial max-pooling across the $K$ dimensions resulting in a $1 \times K$ dimensional feature vector $\mathbf{f}_{R_i}$ per region $R_i$ , where

$$\mathbf{f}_{R_i} = [f_{R_i,1}...f_{R_i,k}...f_{R_i,K}],$$
$$\text{such that } f_{R_i,k} = \max_{p \in R_i} X_{k,p}^l \qquad (1)$$

Each region vector $\mathbf{f}_{R_i}$ is $l_2$ normalized, followed by whitening with PCA and $l_2$ normalized again. The regional feature vectors are sum aggregated to get the final image representation $\mathbf{f}$ :

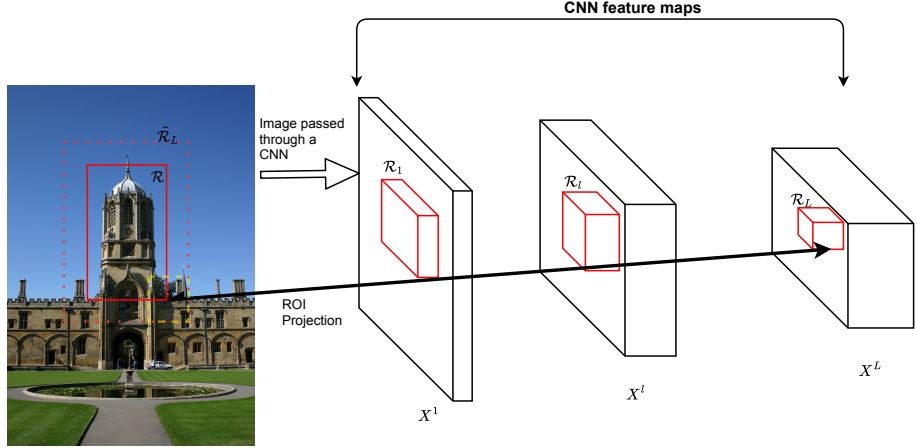

**Fig. 1.** An overview of feature extraction using a CNN. The network has $l$ layers, $l=1..L$, where each layer computes a feature map $\boldsymbol{X}^l$ using the representations from the previous layers. The representations $(\mathcal{R}_1..\mathcal{R}_l..\mathcal{R}_L)$ denote projections of the ROI, $\mathcal{R}$, on feature maps from different layers, $l$ (bold black line)[39]. $\tilde{\mathcal{R}}_L$ is the cumulative receptive field of the activations inside $\mathcal{R}_L$.

$$\mathbf{f} = \sum_{i=1}^{N} \mathbf{f}_{R_i} \qquad (2)$$

As a result of sum aggregation, the final feature vector still retains the $1 \times K$ dimensionality. The final image representation $\mathbf{f}$ is again $l_2$ normalized, such that a simple dot product can be used to compute image similarity.

## 3    Contextual Information

In the general setting of particular object retrieval, contextual information can be viewed as the information outside the ROI, $\mathcal{R}$ in the query image. Such information can be facilitatory or inhibitory in the retrieval process. In other words, the contextual information along with ROI can increase or decrease the distinctiveness of the query image. Traditional models for extracting query image CNN representations include :

**Full Query (FQ)** : The full query image is fed to a CNN [14], and the resulting feature representation is encoded to fixed length feature vector.

**Cropped ROI (RQ)** : The ROI in the query image is cropped [27], [20] and fed into a CNN, followed by encoding the resulting feature map.

**Cropped Activation (AQ)** : The feature map representation is first obtained by feed-forwarding the whole query image through a CNN. Projection of the ROI on the feature map is computed, which is represented by the set of activations that have their center of receptive fields inside the ROI as shown in Fig.1. The representations within the projected ROI are then encoded using standard encoding methods [27], [14], [19].

Methods that use **FQ** models are expected to perform better than **RQ** based methods when the contextual information is facilitatory. However, the facilitatory nature of context almost cannot be known apriori. On the other hand, methods that extract query image representation using **AQ** [27], [14], [19] are able to encode certain amount of contextual information. As can be seen from the Fig.1, the total receptive field, $\tilde{\mathcal{R}}_L$ [40](dotted red colored box) of the activations in the projected ROI, $\mathcal{R}_L$ extend beyond the ROI, $\mathcal{R}$. Thus the extended ROI ($\tilde{\mathcal{R}}_L$ in Fig.1) encodes certain amount of context. Note that the receptive field of layers $l < L$, $\tilde{\mathcal{R}}_l$ will have a smaller area than $\tilde{\mathcal{R}}_L$.

However, due to limited reach of the receptive field of the boundary activations, a large amount of contextual information is discarded. We propose to combine the advantages of the three models mentioned above, by leveraging the computational model of spatial attention observed in the hippocampus to extract full query image CNN representation. The attention model is presented next.

### 3.1   Computational model of Spatial Attention

The main advantage of **RQ** and **AQ** models is that the representation from the ROI has the highest response in the final query representation. Particularly, in **RQ**, ROI representation has the sole representation, while, in **AQ**, it has higher representation than the context. For **AQ**, this is based on the assumption that only boundary activations are affected by regions outside the ROI. On the other hand, the disadvantage of **FQ** based methods is that the ROI ceases to have a higher prominence in the final representation.

Thus, an ideal model should not only encode information from regions beyond the ROI, but, in the process should also maintain higher response from the ROI in the final representation. Such a constraint can be modelled using the computational model of spatial attention observed in hippocampus [38].

The attention model initiates with a saliency or attention map $A \in \mathbb{R}^{W_a \times H_a}$ defined over a feature map $\boldsymbol{X}^l$, such that $W_a = W$ and $H_a = H$. We do not introduce any layer specific notations as the mask can be applied to feature map at the output of any layer, $l$, of a CNN architecture. The same attention mask is applied across all the channels, $K$, of the feature map $\boldsymbol{X}^l$ (see Sec.2.1). Therefore, each element $p$ in the mask $A$, $A_p \in [0,1]$, affects the activations occurring at the spatial location $p$ across the $K$ channels, $X^l_{1:K,p}$. The activity levels are defined as follows :

$$A_p = \begin{cases} 1, & \text{if } p \in \mathcal{R}_l \\ M_p, & \text{if } p \notin \mathcal{R}_l, \end{cases} \tag{3}$$

where $\mathcal{R}_l$ is the projection of the ROI, $\mathcal{R}$, onto the feature map $\boldsymbol{X}^l$ [39] (see Fig.1 ), and, $M$ is the saliency map introduced in Sec.4.1. Note that the attention mask is specific to the spatial location and independent of the channel dimension. So, identifying the activations, across different channels, with their spatial location suffices, i.e. the notation $p \in \mathcal{R}_l$ represents all activations occurring at spatial location $p$ in the feature map $\boldsymbol{X}^l$ and lying within $\mathcal{R}_l$, $X^l_{1:K,p} \in \mathcal{R}_l$. Now, activations occurring at position $p$ in each feature map $X^l_k \in \boldsymbol{X}^l, k = 1...K$, are modulated by the attention mask as follows :

$$\tilde{X}^l_{k,p} = \begin{cases} A_p X^l_{k,p}, & \text{if } p \in \mathcal{R}_l \\ g(A_p) X^l_{k,p}, & \text{if } p \notin \mathcal{R}_l, \end{cases} \qquad (4)$$

$g(.)$ is a monotonic function [38] :

$$g(a) = \lambda_1 + \lambda_2 a^\phi \qquad (5)$$

The constants $\lambda_1, \lambda_2 \in (0, 1)$ are so chosen such that the function $g(.)$ always maintains a value less than one i.e. $g(.) < 1$. Additionally, the function $g(.)$ has a lower bound at $\lambda_1$, that defines the maximum attenuation which can be applied to any activation outside the projected ROI, $\mathcal{R}_l$. In all our experiments we set $\lambda_1 = 0.5$ , and,$\lambda_2 = 0.4$. The constant $\phi$ suppresses activations with weak attention levels (less salient). As in [38], we set $\phi = 4$ for all experiments.

The modulated feature map representations $\tilde{\boldsymbol{X}}$ are feed-forwarded through the remaining layers of the CNN network, or, directly encoded into a fixed length global image representation (detailed in Sec.5).

For theoretical purposes, we would like to discuss the related work of Chum et al. [4] in the light of spatial attention model. Chum's model (**Ctx**) first detects affine invariant Hessian regions (features) in the whole query image. The regions inside the ROI are identified as active (attention level = 1) and those outside as inactive (attention level = 0). The active regions, described by SIFT descriptors are then used to query the database. The top ranked images from the initial retrieval are then passed through a costly spatial re-ranking step (geometric verification). The inactive features in the query image are activated by features from top-ranked database images with same visual word and similar geometry. The expanded query model is then used to query the database. Thus the **Ctx** model incurs additional cost in terms of querying the database multiple times and costly spatial re-ranking steps to learn relevant contextual information. Instead, context aware CNN representations outputted by **SA** model can be encoded into a single global image representation, requiring only a single pass through the database to obtain better retrieval performance than context devoid representations.

## 4    Weighted R-MAC

The feature map is now encoded to a lower dimensional feature representation. Recent state-of-the-art use R-MAC for this operation [27], [20]. However, one of the criticality with the standard R-MAC is the equal weighting of each region vector $\mathbf{f}_{R_i}$ while aggregating in equation 2. This implies that responses generating from image clutter and background, will negatively affect the retrieval process due to assignment of uniform weightage. As one does not have the prior information about the location of the object of interest in the database images, increasing the number of regions ensures higher coverage, but, it also increases interference from irrelevant regions.

Instead, we propose to use a weighted version of the standard R-MAC (WR-MAC) such that the final image representation is a weighted combination of representations from each region $R_i \in \mathbf{R}$. In general, equation 2 is modified to

$$\mathbf{f} = \sum_{i=1}^{N} w_i \mathbf{f}_{R_i} \qquad (6)$$

The weights $w_i \in \mathbb{R}$ are generated using a saliency measure (Sec.4.1) such that landmark type regions have higher weights than background clutter.

### 4.1   Saliency Map

We use a simple, yet effective saliency measure [31], [32]. Given the feature map $\mathbf{X}^l$, the saliency function maps the 3D tensor $\mathbf{X}^l$ to a 2D tensor $M$ by sum aggregating the feature map $\mathbf{X}^l$ over the channel dimensions. Mathematically, the function can be defined as $\psi : \mathbb{R}^{W \times H \times K} \to \mathbb{R}^{W \times H}$ such that $\psi(\mathbf{X}^l) = M$, where $M = \sum_{k=1}^{K} X_k^l$. The map is additionally max-normalized such that each element $p$ has a range, $M_p \in [0, 1]$.

In Sec.2.2 where we introduced R-MAC, we observe that the regions $\mathbf{R}$ generated by the rigid grid depend on the spatial dimension $\mathbf{S}$ of the feature map $\mathbf{X}^l$. As the spatial dimension of feature map $\mathbf{X}^l$ is retained in the saliency map $M$, we can define the same set of regions $\mathbf{R}$ over the map $M$. Now, for each region $R_i$ we compute MAC to obtain the weight $w_i$. That is

$$w_i = \max_{p \in R_i} M_p \qquad (7)$$

## 5   Experiments

### 5.1   Network and Datasets

For all our experiments we use the ResNet-101 [30] CNN network architecture. The network has a very deep architecture and attains the state-of-the-art results in a variety of computer vision problems [30]. The original network, pre-trained on ImageNet, is cropped at the layer `res5c_ReLU` and fine-tuned using the method proposed in [37]. The network parameters of the fine-tuned model (ResNet-IR) are also publicly available [41]. Additionally, the model ResNet-IR has additional layers on top of `res5c_ReLU` which performs PCA whitening on the region vectors obtained from `res5c_ReLU` representations using R-MAC grid and performs the final R-MAC aggregations. The additional layers were trained

**Table 1.** Effect of applying attention model to different CNN layers, measured in terms of mAP

| Layers | Paris6k | | Oxford5k | |
|---|---|---|---|---|
| | R-MAC | WRMAC | R-MAC | WRMAC |
| `res2c_ReLU` | 95.44 | 95.68 | 87.90 | 88.49 |
| `res4b15_ReLU` | 95.49 | 95.80 | 89.45 | 90.20 |
| `res5c_ReLU` | 95.53 | 95.84 | 89.04 | 89.37 |

**Table 2.** Performance comparison of our proposed model of extracting query image representation (Spatial Attention, **SA**) and encoding using weighted R-MAC (WR-MAC), with existing approaches. Previous baseline [37] is marked as [†]. Re-ranking stages were not applied.

| Models | Paris6k | | Oxford5k | |
|---|---|---|---|---|
| | R-MAC | WRMAC | R-MAC | WRMAC |
| Cropped ROI (**RQ**) | 94.5[†] | 94.8 | 86.1[†] | 86.7 |
| Cropped Activation (**AQ**) | 94.81 | 95.13 | 88.77 | 89.03 |
| Full Query (**FQ**) | 95.47 | 95.76 | 87.30 | 87.86 |
| Spatial Attention (**SA**) | 95.49 | **95.80** | 89.45 | **90.20** |

end-to-end during fine-tuning [37]. We use the learnt PCA parameters from [37] in our experiments.

For evaluation, we use the standard and publicly available particular object retrieval datasets : Oxford Buildings [3] and Paris Buildings [21]. Each dataset contains several thousands of images. Within these images, 55 queries are defined and a ROI containing the precise location of the landmark to be queried is defined. The retrieval performance is computed using mean average precision (mAP), where the mean is taken over all queries. Re-ranking operations like query expansion [4] and database augmentation [37] were not applied in the presented results.

### 5.2 WRMAC

As mentioned in Sec.2.2, for a given image, R-MAC can be used to encode representations from any given CNN layer, but, similar to [37] we use the `res5c_ReLU` representations for WRMAC. A saliency map is computed using the `res5c_ReLU` representations which is then normalized. The normalized map is used to generate weights. The weights are only applied at the time of aggregating the regional representations generated from the R-MAC grid. Note that prior to aggregation, these representations are $l_2$ normalized, whitened with PCA and $l_2$ normalized again. Similar to [37], we extract R-MAC and WRMAC representations from 3 different scales of the given image: 550, 800, and, 1050 pixels on the largest side and maintaining the original aspect ratio. The representations from the 3 scales are aggregated and $l_2$ normalized to obtain the final image representation.

### 5.3   Attention model

The only parameter in the attention model which needs to be defined is the layer number, $l$, i.e. the CNN layer on which the attention model is to be applied. As at each layer, feature representations are computed using the representations from the previous layers, application of attention model at a certain layer affects the representations of all the layers above. As the number of layers is high for ResNet-IR network, we choose the following layers: i) `res2c_ReLU`, ii) : `res4b15_ReLU`, and, iii) : `res5c_ReLU` and evaluate their performance as described below.

Given a query image and a ResNet-IR layer, $l'$ (sampled as mentioned above), the feature representations at the output of layer $l'$ are used to compute the normalized saliency map. Thus, it is to be noted that the saliency map used in the experiments vary, i.e. for WRMAC the saliency map is defined over the `res5c_ReLU` representations, while for attention model it is defined over the layer, $l'$. The modulated representations of the layer, $l'$ are the used as an input to the layer, $l' + 1$, and, feed-forwarded through the remaining layers of ResNet-IR to obtain the final `res5c_ReLU` representations. The `res5c_ReLU` representations are then encoded as discussed in Sec.5.2.

### 5.4   Results

In order to better interpret the results, the following points need to be noted : i) the spatial attention model is only applied to the query side, and, ii) the proposed encoding method, WRMAC, is applied to both the query and database images.

The motivation for evaluating the performance of attention model across different CNN layers is to observe where in the CNN should the contextual representations be suppressed. In other words, should it be suppressed as early as lower level layers (`res2c_ReLU`), or, late at high level layers (`res5c_ReLU`), or, somewhere in between at mid level (`res4b15_ReLU`). From the results in Table 1, it can be observed that performance across all the layers (considered in Table 1) improve over the baseline (Table 2). However, the performance of `res4b15_ReLU` layer representations is more consistent across the datasets. As such we consider spatial attention model (**SA**) applied on `res4b15_ReLU` layer representations as our proposed model and compare with the existing approaches in Table 2. Similar to Sec.5.3, for all the models in Table 2, the `res5c_ReLU` representations are obtained and then encoded as discussed in Sec. 5.2. All the models **FQ**, **AQ**, **SA** perform better than the baseline **RQ** due to the addition of contextual information in the final image representation. However, the **SA** model outperforms the existing models across both the datasets.

The comparison between the performance of **FQ** and **AQ** models gives an idea of the effect of the contextual information in image retrieval. The model **AQ**, which encodes limited contextual information, outperforms **FQ** in Oxford dataset, but, is outperformed by **FQ** in Paris dataset. Our assumption is that in Paris dataset, the contextual information has a facilitatory effect, while, in Oxford dataset, it has a certain degree of inhibitory effect on the retrieval performance. An in-depth analysis is left for future work. Although, the performance of **SA** model is constrained by the requirement of ROI on the query side, that does not diminish the scope of the proposed encoding method, which shows consistent improvement even without the spatial attention model.

## 6 Spatial Attention on Database

As the **SA** model requires a ROI to attend to, its functionality remains restricted to only query images. In order to remove this asymmetric functionality, we propose a simple extension of the model to database images in this section.

The motivation for applying **SA** model on query images as presented in Sec. 3 also extends to database images. Thus, similar to the procedure described in Sec. 5, we apply **SA** on `res4b15_ReLU` layer representations of database images. But first the projection of the potential ROI on `res4b15_ReLU` layer representations are required which is obtained as follows. In the first feed-forward pass through the CNN, we extract representations from the `res5c_ReLU` layer. The representations are then used to obtain the normalized saliency map and additionally encoded using WRMAC (the requirement of which is explained later in the section). A binary map is computed by thresholding the saliency map at the level, $\tau = 0.7$ such that elements of the map having saliency greater than $\tau$ are assigned a value of 1, while the rest are assigned 0. The binary map is then resized to match the spatial resolution of the `res4b15_ReLU` layer representations. Using connected component clustering on the binary map, multiple bounding boxes are generated which can be defined as the projections of potential ROIs on `res4b15_ReLU` layer representations in database images.

Now for each ROI, the **SA** is applied on `res4b15_ReLU` layer representations that are then feed-forwarded to obtain the `res5c_ReLU` layer representations. Finally, the representations are encoded using WRMAC. The process is repeated independently for each ROI, and in the end WRMAC representations from all the ROIs and that obtained from the first pass (mentioned above) are aggregated and $l_2$ normalized. The WRMAC representation from the initial pass is used as sometimes in a database image the expected ROI specified by the query image may not always have saliency higher than the threshold, $\tau$. In such cases, the suppression effect from **SA** will only negatively affect the retrieval performance.

### 6.1 Evaluation

Table 3 shows the performance evaluation of **SA** applied on query images as compared to its application on both query and database images. It shows that using **SA** model on the database side improves the retrieval performance. Although the performance increase is not large, the simplicity of the proposed approach leaves room for future work.

## 7 Conclusion

In this paper we address the different models of extracting and encoding image representation using CNN for the task of image retrieval. The current mod-els of extracting CNN representations from query images cannot fully exploit the advantages of the provided ROI in the query image. We propose an extension of an attention model to this problem and demonstrate increase in retrieval accuracy on standard particular object retrieval datasets. Additionally, we also propose an extension to a recently introduced popular encoding method for CNN representations, R-MAC and show that together with the attention model, the new encoding strategy achieves state-of-the-art accuracy on object retrieval benchmarks. To the best of our knowledge our results are the best among global image representations reported so far for these datasets. Using efficient re-ranking strategies [4], [42] can lead to further improvements in retrieval accuracy. The code for the proposed model is available at `https://github.com/AaltoVision/Object-Retrieval`.

**Table 3.** Performance evaluation of Spatial Attention model (**SA**) when applied on just the query images ($SA_q$) and both query and database images ($SA_{q\_dB}$).

| Models | Paris6k WRMAC | Oxford5k WRMAC |
|---|---|---|
| $SA_q$ | 95.80 | 90.20 |
| $SA_{q\_dB}$ | **95.97** | **90.45** |